\documentclass[11pt]{article}

\usepackage[preprint]{acl}
\usepackage{float}
\usepackage{times}
\usepackage{latexsym}
\usepackage{xcolor}
\usepackage{expex}
\usepackage[T1]{fontenc}

\usepackage[utf8]{inputenc}

\usepackage{microtype}

\usepackage{inconsolata}

\usepackage{graphicx}

\title{GlossAssist - A Tool to Simplify Corpus Creation and Study the Effect of NLP Models in Low-Resource Documentation Settings}

\author{
  Bhargav Shandilya \And
  Matt Buchholz \And
  Alexis Palmer \AND
  University of Colorado Boulder \\
  \texttt{\{bhargav.shandilya, Matthew.Buchholz, alexis.palmer\}@colorado.edu}
}
\begin{document}
\maketitle
\begin{abstract}

Interlinear glossed text (IGT) is the standard format for linguistic annotation in language documentation. Producing it manually, however, is often slow and costly. Automated glossing systems have improved substantially in recent years, but adoption among field linguists remains limited \cite{rice-etal-2025-interdisciplinary}. Existing tools are designed to be evaluated rather than used, offering no interpretable path for correction or the incorporation of linguistic expertise back into model behavior. We present GlossAssist, a glossing tool built around the retrieval-based architecture of CWoMP (Contrastive Word-Morpheme Pre-training) \cite{alper2026cwompmorphemerepresentationlearning}, which grounds predictions in a mutable lexicon of learned morpheme representations. In conjunction with CWoMP, our system treats each correction by an annotator as part of an active learning setting, which expands the lexicon and improves future predictions without having to retrain the model. In this paper, we present our interface and argue that this feedback loop should be treated as a design requirement for NLP tools aimed at documentary linguists.

\end{abstract}
\begin{figure}
    \centering
    \includegraphics[width=0.83\linewidth]{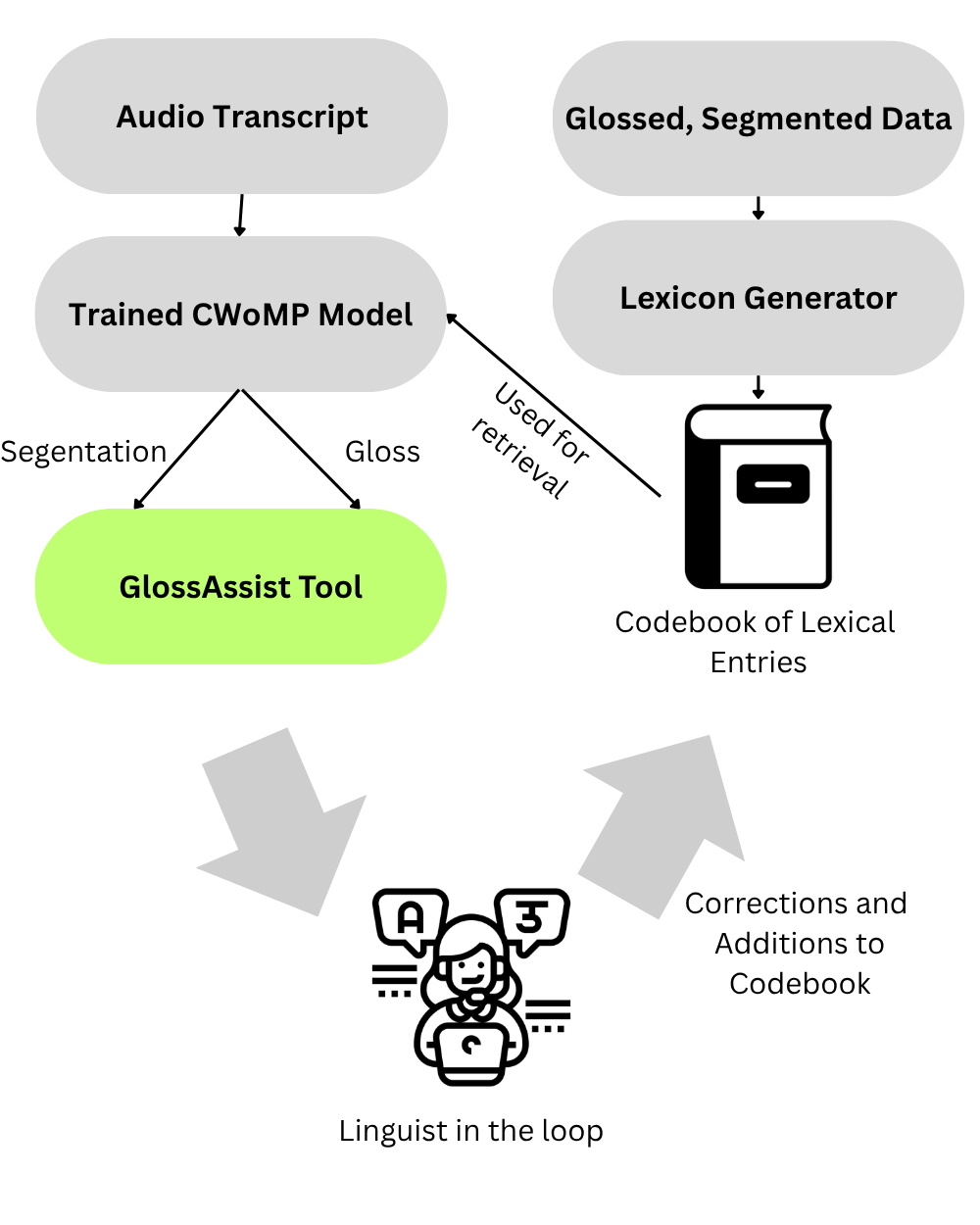}
    \caption{GlossAssist Workflow with CWoMP Integration}
    \label{fig:linguist_in_the_loop}
\end{figure}
\section{Introduction}

A majority of the world’s languages spoken today are endangered or close to extinction, and the linguistic knowledge encoded in their morphological systems, grammatical structures, and oral tradition cannot be replaced once lost. Interlinear Glossed Text (IGT) is a standard approach in the field to capture this linguistic knowledge. The process of annotating a corpus in IGT format involves extensive linguistic analysis, including morphological segmentation and morpheme glossing (See Appendix \ref{app:igt} for more details).

The computational linguistics community has contributed several automated glossing systems \cite{girrbach-2023-tu-cl,ginn2024glosslmmassivelymultilingualcorpus,ginn2026massivelymultilingualjointsegmentation,shandilya-palmer-2025-boosting, coltekin-2019-cross, wiemerslage-etal-2023-investigation}. Recent models have achieved impressive benchmark performance across typologically diverse languages on tasks such as segmentation, glossing, and translation. Adoption among documentary linguists, however, remains low. The problem goes beyond just raw accuracy and model hallucinations. It is that existing systems are designed more for evaluation than practical use. They produce outputs that are difficult to inspect, correct, or trust, treating the linguist as a passive user of model predictions rather than an expert collaborator. Moreover, it is difficult for linguists to switch over to a new tool unless the tool provides very strong incentives for use. While the advances in language models have been significant, many of them are locked away behind hard-to-use code repositories that are inaccessible to researchers who are unfamiliar with modifying and running code. Several of these repositories are also not maintained, making it difficult to reproduce results or actively use the code to conduct further investigations.

We present a glossing tool designed around a different set of assumptions, treating the linguist as an important part of model refinement. The system is based on the retrieval-based architecture of CWoMP \cite{alper2026cwompmorphemerepresentationlearning}, which grounds predictions in a mutable lexicon of learned morpheme representations. We position this feedback loop as a way to go beyond benchmarking and towards a framework that is measurably usable in the field.

\section{Related Work}

\paragraph{Automated Glossing.}Early approaches to automated IGT generation used rule-based and statistical methods \cite{evaluating-automation,moeller-hulden-2018-automatic}, while the SIGMORPHON 2023 shared task \cite{ginn-etal-2023-findings} introduced several neural approaches \cite{girrbach-2023-tu-cl, cross-etal-2023-glossy} . Most of these treat glossing as a character- or byte-level sequence-to-sequence task, with Polygloss \cite{ginn2026massivelymultilingualjointsegmentation} representing the current state of the art through multilingual pretraining on a large cross-lingual IGT corpus. Recent work has also explored in-context learning with LLMs \cite{ginn-etal-2024-teach}, finding that current models struggle with endangered languages despite strong multilingual performance elsewhere. \cite{zhao-etal-2020-automatic} demonstrate the value of leveraging translations as an additional input signal, a design choice our tool inherits from the CWoMP system. In contrast to these systems, which generate glosses as unconstrained text, both CWoMP and our tool treat morphemes as atomic form-meaning units grounded in a retrievable lexicon, thus preventing hallucination by design and enabling inference-time extension without retraining.

\paragraph{Computational Methods for Language Documentation.} A parallel body of work addresses the broader challenge of integrating computational tools into documentary linguistics workflows. \citet{gessler-2022-closing} diagnose the infrastructure gap directly, arguing that NLP and documentary linguistics need shared software infrastructure to bridge the disconnect between research systems and field practice. \citet{moeller-arppe-2024-machine} describe a curriculum for teaching documentary linguists to adopt a machine-in-the-loop approach, observing that non-computational linguists tend to prefer cleaning existing data over expanding training sets, a finding that directly motivates our tool's lexicon-expansion design over a retraining paradigm. \citet{rice-etal-2025-interdisciplinary} provide possibly the most detailed account of what documentary linguists actually need from automated glossing tools, identifying misaligned segment-gloss pairs and hallucinated morpheme types as the most damaging failure modes in practice. Our tool is designed around both of these findings. The ComputEL workshop series \cite{computel-ws-2025-main} more broadly represents the community working at this intersection and provides important context for understanding the practical constraints under which field linguists operate.

\begin{figure*}
    \centering
    \includegraphics[width=1\linewidth]{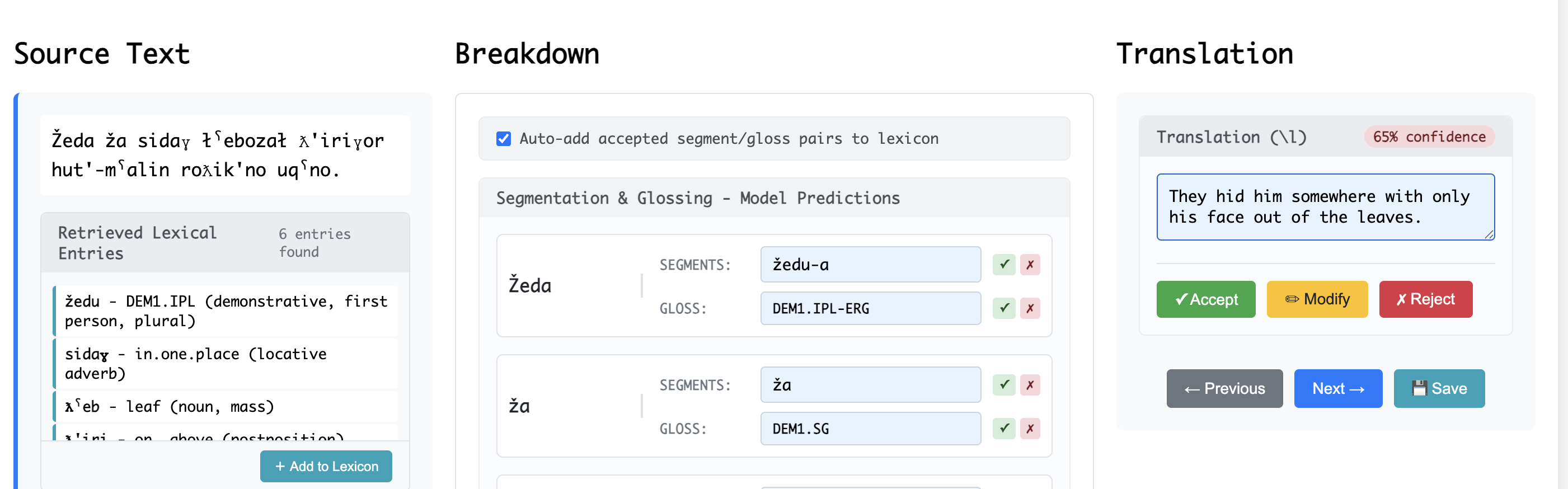}
    \caption{GlossAssist Annotation Interface}
    \label{fig:annotation_interface}
\end{figure*}

\paragraph{Annotation Tools and Linguistic Infrastructure.} Field linguists use a wide range of tools and workflows, many of them relying on ELAN\footnote{\url{https://archive.mpi.nl/tla/elan}} for time-aligned transcription and FLEx\footnote{\url{https://software.sil.org/fieldworks/}} for morphological analysis and lexicon management, with established workflows integrating the two. ELAN's Lexan \footnote{\url{https://pure.mpg.de/view/item_1838141}} framework offers some support for plugging in computational annotation modules, but as noted above, seamless NLP integration for low-resource languages continues to be elusive.
Our tool complements these platforms, focusing specifically on the IGT annotation bottleneck and the feedback loop between model predictions and lexicon growth.

\paragraph{Human-in-the-Loop Annotation.} The broader NLP literature on human-in-the-loop annotation is relevant to our position on workflow-centered evaluation. \citet{Settles2009ActiveLL} surveys the literature that establishes active learning as a framework for prioritizing human annotation effort. Subsequent work has further explored how model-assisted annotation potentially alters annotator behavior, sometimes in undesirable ways. For example, \citet{schroeder-etal-2025-just} show that LLM-generated suggestions do not necessarily improve human annotation speed or accuracy. Our tool's design takes this risk into account by showing predictions as hypotheses to be evaluated at the morpheme level. The logged feedback data also enables the kind of behavioral analysis that \citet{schroeder-etal-2025-just} call for, namely tracking how and when annotators override model outputs.

\section{GlossAssist Architecture}

Our tool\footnote{https://github.com/bhargavns/GlossAssist} consists of two integrated components: \textbf{(1) an annotation interface} for sentence-by-sentence IGT production, and \textbf{(2) a researcher dashboard} for analyzing model performance across annotation sessions. Together, they are designed to support three modes of use - integrating existing NLP models into active documentation workflows, evaluating model efficacy on a corpus, and building corpora from scratch.
The annotation interface is organized as a three-panel layout, as shown in Figure~\ref{fig:annotation_interface}. The left panel displays the source text for the current sentence alongside two automatically populated reference boxes: retrieved lexical entries matching morphemes in the input, and retrieved grammatical rules relevant to the sentence's morphosyntax. These are drawn from the growing lexicon and presented to the linguist as transparent evidence for the model's predictions, rather than staying hidden within a black-box output. The center panel is the primary workspace. It presents CWoMP's joint segmentation and glossing predictions word by word. Each word is displayed alongside editable fields for its predicted segment string and gloss, with per-prediction accept and reject controls. Crucially, accepted segment/gloss pairs are automatically added to the mutable lexicon via a toggleable auto-add mechanism. The right panel handles translation: a model-predicted translation is displayed with a confidence score, and the linguist can accept, modify, or reject it independently of the morpheme-level decisions \footnote{CWoMP does not produce the translation as an output. Any suitable translation model can be used in conjunction with GlossAssist}. Navigation controls allow movement between sentences, and the interface logs a timestamped record of every accept, reject, and modification event.

This event log feeds the second component: a researcher dashboard (See figure \ref{fig:user_dash}) that aggregates feedback data across sessions. Session-level KPIs such as average time per sentence, overall prediction accuracy, and total feedback events give an immediate quantitative overview of annotation time and model reliability. The dashboard thus closes a second feedback loop between annotation patterns and the researcher's understanding of where model improvement is most needed.

Taken together, these components reframe the annotation session where accepts are validated lexicon entries, rejects are a signal about model failure, and every modification is a gold-standard correction. GlossAssist is designed to make this structure explicit and actionable, for both the linguist annotating in the field and the researcher evaluating model behavior across a corpus.

\textbf{A note on error propagation} - Error propagation is a concern with a tool like GlossAssist. Model predictions can be compelling at times, and users might be tempted to accept the predictions without proper scrutiny in extended annotation sessions under time pressure \cite{schroeder-etal-2025-just}. We are working to build in visual indicators for model confidence and likely out-of-vocabulary(OOV) candidates to address this problem.

\section{Position}
We argue that the design of GlossAssist embodies three major claims about what NLP systems for language documentation should look like:

\paragraph{Annotation effort does not need to be divorced from model refinement.} The standard framing of automated glossing treats human annotation as a cost to be minimized: the goal is to reduce the number of decisions a linguist has to make. We, instead, propose that each decision made by a linguist is an investment that makes the system more capable. In GlossAssist, accepted predictions are absorbed into the lexicon and retrieved for future sentences. The result is a compounding efficiency curve. Naturally, early annotation sessions require more correction compared to later sessions. As the lexicon grows, the model's coverage improves, and the proportion of predictions requiring intervention reduces. This closely resembles the natural workflow of a field linguist building up a language description over time, and it aligns model improvement with the linguist's existing practice rather than requiring a separate technical intervention like retraining. \citet{alper2026cwompmorphemerepresentationlearning} provide direct empirical support for this claim. The mutable lexicon evaluation in CWoMP shows that, across languages, extending the lexicon with ground-truth morphemes consistently reduces Morpheme Error Rate (MER), with the largest gains in the lowest-resource settings where compounding returns matter most.

\paragraph{Interpretability is a prerequisite for trusting model outputs.} A core finding of \citet{rice-etal-2025-interdisciplinary} is that field linguists have been reluctant to adopt automated glossing tools because the tools' outputs are difficult to inspect and correct. When a model produces a hallucinated morpheme type or a misaligned segment-gloss pair, a linguist working on an endangered language has no safe way to address it at the model level. There's also no mechanism for understanding why it occurred. GlossAssist addresses this directly. Because CWoMP's predictions are grounded in a discrete lexicon of verified morpheme entries, every output is traceable to a specific lexical decision. The retrieved lexical entries and grammatical rules displayed in the source panel make this grounding visible to the user. Hallucination of unattested morpheme types is prevented by design, and segment-gloss alignment is guaranteed structurally. The linguist is shown the evidence and asked to evaluate it, instead of being presented with a black box.

\paragraph{Evaluation should not be entirely benchmark-centered.} Current evaluation of automated glossing systems relies almost exclusively on corpus-level accuracy metrics such as MER, WER, and chRF scores measured for isolated test sets. These metrics are valuable for comparing systems, but they are poorly matched to the concerns of a documentary linguist - annotation speed, model adaptability, ease-of-use, and interpretable outputs. The GlossAssist dashboard is designed to make these questions answerable. Logging every annotator decision with a timestamp and morpheme index produces the data needed for workflow-centered evaluation. We argue that this framing of evaluation, grounded in annotation efficiency and error diagnostics rather than held-out accuracy alone, should become standard practice for NLP tools aimed at documentary linguists.

\section{Conclusion}
Language documentation is one of the few scientific endeavors in which the window for collecting low-resource language data is closing rapidly. Automated glossing research has made genuine progress on benchmark accuracy, but progress on adoption has lagged, and we argue that this gap is structural rather than incidental.
By grounding predictions in an inspectable, extensible lexicon, showing evidence alongside outputs, and treating every annotator decision as structured data, our tool allows linguists to become active collaborators whose expertise improves the system over time. 
Using GlossAssist, we plan to conduct user studies with practicing documentary linguists, measuring annotation efficiency and the effect of active learning across several low-resource languages. Beyond evaluation, future work should explore integration with existing tools such as ELAN and FLEx, multi-annotator workflows for collaborative corpus projects, and extension to speech input for settings where transcription and glossing must happen simultaneously. 
Ultimately, we hope this work contributes both a tool and a fresh framing of the relationship between NLP systems and documentary linguists working to preserve languages.

\section{Ethical Considerations}

We are aware of the ethical considerations raised by work on languages spoken by Indigenous communities. Tools built by researchers outside low-resource language communities always run the risk of carrying cultural biases. 
Members of language communities should be able to directly benefit from any research done on their language. 

One of the goals of GlossAssist is to lower the barrier to entry for producing interlinear glosses for a wide range of languages by making the annotation process faster and easier. By reducing the burden of annotation, we aim to make technologies and materials based on these glosses more reliable and easier to adopt. 

\section{Limitations}

Because CWoMP serves as the backing prediction model for GlossAssist, the system's predictions are limited to a discrete morpheme lexicon. Therefore, the system cannot currently generate glosses for morphemes absent from the lexicon, though we are actively working to integrate out-of-vocabulary (OOV) prediction capabilities. Especially in early stages of documentation, this lack of OOV prediction capability means greater effort is required of the annotator to expand the lexicon and improve coverage of the CWoMP model.


\bibliography{custom}
\section*{Appendix}
\appendix

\section{Interlinear Glossed Text}
\label{app:igt}
Interlinear Glossed Text (IGT) is a standard format for morpheme-level linguistic annotation in the field of language documentation. An IGT instance will typically contain the \textbf{original transcription}, a \textbf{morphological segmentation} tier, a \textbf{gloss tier}, and a \textbf{free translation}. These conventions follow the Leipzig Glossing Rules \cite{Leipzig}. The following examples illustrate the format:

\begin{small}
\ex 
\begingl
\gla xqil//
\glb x-$\emptyset$-q-il//
\glc COM-A3S-E1P-ver//
\glft `lo vimos' (`we saw it')//
\endgl
\xe

\ex 
\begingl
\gla Ražbadinez idu//
\glb Ražbadin-z idu//
\glc Razhbadin-GEN2 home//
\glft 'Razhbadin's home'//
\endgl
\xe

\end{small}

\textbf{Segmentation tier} - Breaks words at morpheme boundaries by marking them with hyphens. Both surface and canonical forms of segmentation are used in practice. Surface segmentation preserves the form as written in the transcription (for example, leave-s for the English plural form of 'leaf') while canonical segmentation presents the underlying form of the word (leaf-s for the plural of 'leaf'). CWoMP works with both types of segmentation.

\textbf{Gloss tier} - Provides a label for each individual segment. Grammatical labels appear as uppercase letters (ERG for ergative) while stem translations (lexical meanings) appear in lowercase. Example 2 clearly shows this distinction.

\textbf{Translation} - Free translation in a high-resource language like English or Spanish.

IGT serves two main purposes - (1) as a documentary record of a language's morphological and syntactic features, and (2) in the case of NLP systems, as a resource for downstream NLP tasks like segmentation, glossing, and translation.

\section{User Dashboard and Log Page}
\label{app:user_dash}
\begin{figure}[H]
    \centering
    \includegraphics[width=1.0\linewidth]{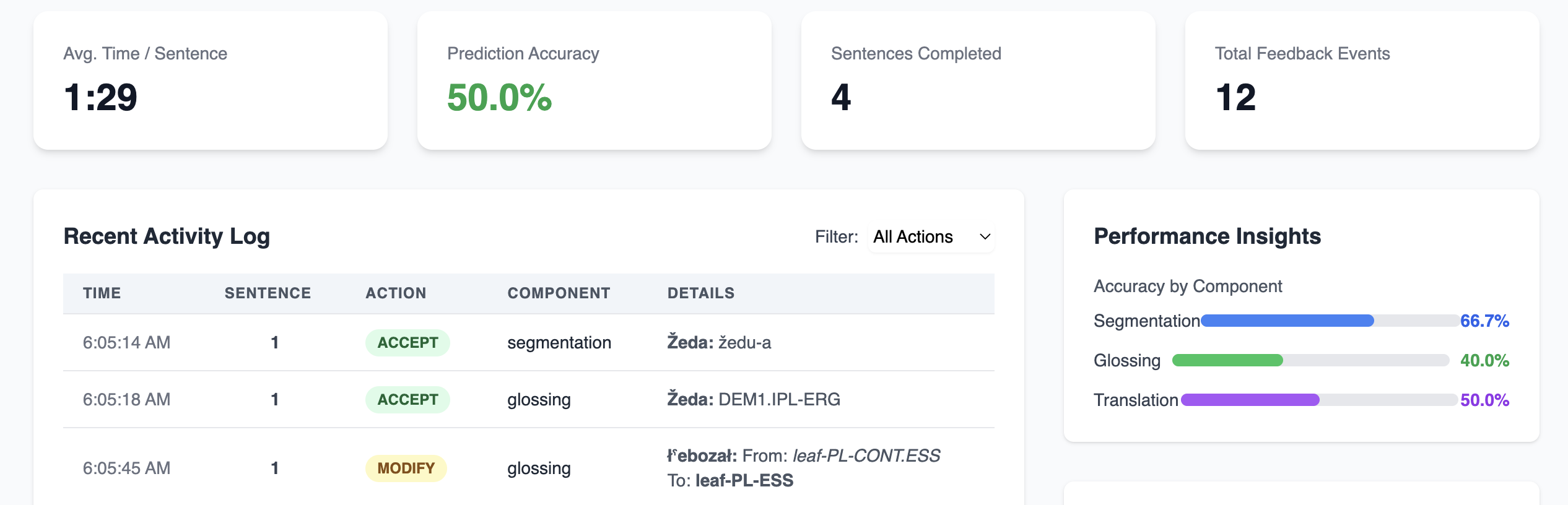}
    \caption{GlossAssist Annotation Interface - User Dashboard and Log}
    \label{fig:user_dash}
\end{figure}

The researcher-facing dashboard contains a structured event log produced by the annotation interface across a session. It provides session-level statistics and fine-grained records of annotator decisions.

The right sidebar provides an additional feature. The user has access to a component-wise accuracy breakdown that distinguishes between segmentation, glossing, and translation performance using labeled progress bars. This allows for the identification of systematic weaknesses based on the annotation tier.

\end{document}